\documentclass[review,11pt]{ReportTemplate}
\usepackage{bm}
\usepackage{graphicx}
\usepackage{paralist,algorithmic,algorithm}
\usepackage{amsmath,mathrsfs,amssymb}
\usepackage{subfigure}
\usepackage{geometry}
\begin{document}
\begin{frontmatter}
\title{Open-environment Machine Learning}

\author{Zhi-Hua Zhou}
\address{National Key Laboratory for Novel Software Technology\\ Nanjing University, Nanjing 210023, China\\
\rm{zhouzh@nju.edu.cn}}

\begin{abstract} 
Conventional machine learning studies generally assume \textit{close-environment} scenarios where important factors of the learning process hold invariant. With the great success of machine learning, nowadays, more and more practical tasks, particularly those involving \textit{open-environment} scenarios where important factors are subject to change, called \textit{open-environment machine learning} (Open ML) in this article, are present to the community. Evidently it is a grand challenge for machine learning turning from close environment to open environment. It becomes even more challenging since, in various big data tasks, data are usually accumulated with time, like \textit{streams}, while it is hard to train the machine learning model after collecting all data as in conventional studies. This article briefly introduces some advances in this line of research, focusing on techniques concerning emerging new classes, decremental/incremental features, changing data distributions, varied learning objectives, and discusses some theoretical issues.
\end{abstract} 
\end{frontmatter}

\section{Introduction}\label{sec:intro}

Machine learning has achieved great success in various applications, particularly in \textit{supervised learning} tasks such as classification and regression. In machine learning, typically, a predictive model optimizing a specific objective is learned from a training data set composed of training examples each corresponding to an event/object. A training example consists of two parts: a feature vector (or called \textit{instance}) describing appearance of the event/object, and a \textit{label} indicating the corresponding ground-truth output. In classification, the label indicates the class to which the training instance belongs; in regression, the label is a real-value response corresponding to the instance. This article mainly focuses on classification, though most discussions are also applicable to regression and other machine learning tasks. Formally, consider the task of learning $f: \mathcal{X} \mapsto \mathcal{Y}$ from a training data set $D = \{(\bm{x}_1, y_1), \ldots, (\bm{x}_m, y_m)\}$, where $\bm{x}_i \in \mathcal{X}$ is a feature vector in the feature space $\mathcal{X}$, and $y_i \in \mathcal{Y}$ is a ground-truth label in the given label set $\mathcal{Y}$.

\begin{figure}[!ht]
\begin{center}
  \includegraphics[width=.6\linewidth]{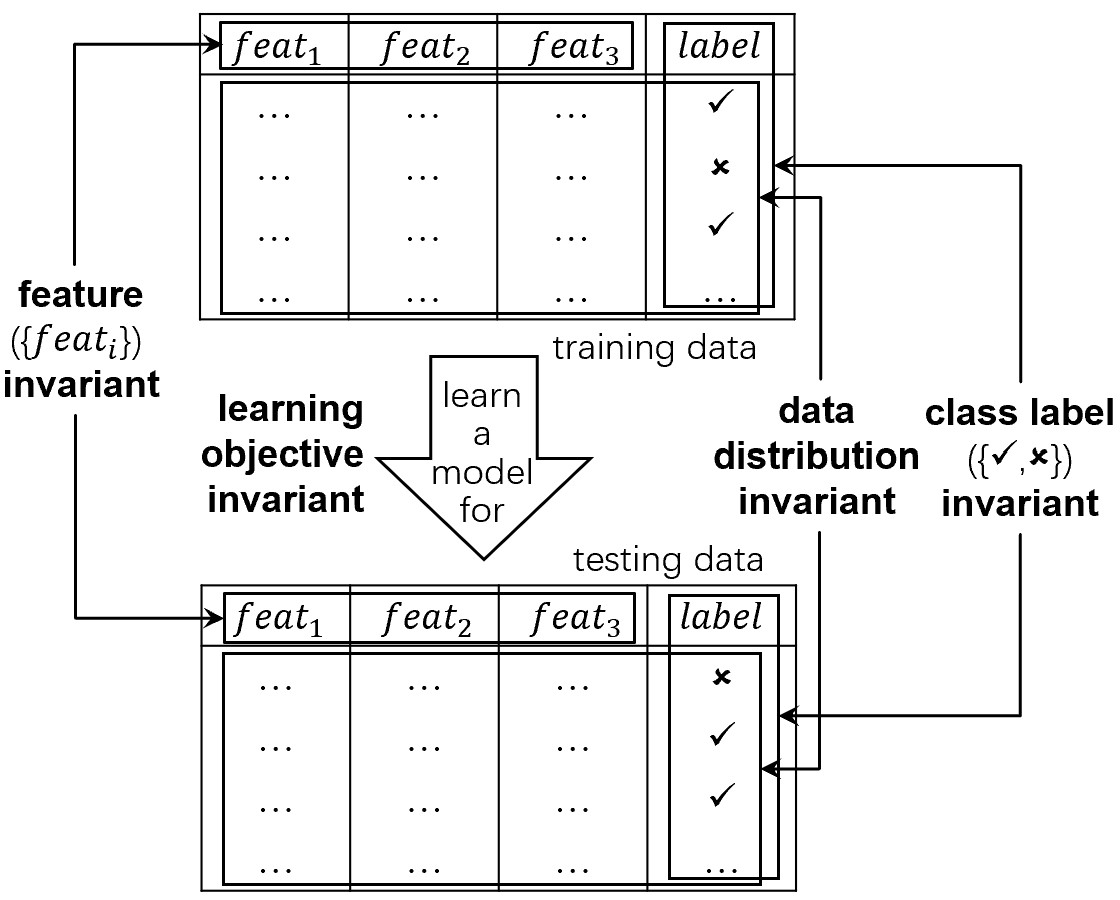}
  \caption{Typical invariant factors assumed in close-environment machine learning.}\label{fig:fig1}
\end{center}
\end{figure}

It is noticeable that current success of machine learning are mostly on tasks assuming \textit{close-environment} scenarios, where important factors of the learning process hold invariant. For example, all the class labels to be predicted are known in advance, the features describing training/testing data never change, all data are from an identical distribution, and the learning process is optimized towards an unchangeable unique objective. Figure~\ref{fig:fig1} illustrates those typical invariant factors assumed in close-environment machine learning studies.

The close-environment assumptions offer a simplified abstraction that enables complicated tasks to be handled in an easier way, leading to the prosperous development of machine learning techniques. With the great achievements attained by these techniques, nowadays, more and more challenging tasks beyond the close-environment setting are present to the community, requesting new generation of machine learning techniques that are able to handle \textit{changes} in important factors of the learning process. We call this \textit{Open-environment Machine Learning} \footnote{The name ``Open-world Machine Learning'' was used to refer machine learning with unseen class \cite{Parmar:Chouhan:Raychoudhury2021} or out-of-distribution (OOD) data \cite{Sehwag:Bhagoji:Song2019}. In fact, it is not beyond closed-environment studies if the unseen class is \textit{known} in advance, and related to Section~\ref{sec:newclass} if unseen class is \textit{unknown}. OOD is related to Section~\ref{sec:changeDistribution}, though concerning only a different distribution is simpler than distribution changing with time.}, or briefly, \textit{Open~Learning} or \textit{Open~ML}.

There seems a straightforward solution: to artificially generate many training examples by mimicking the possible changes in advance, and then feed these data to a powerful machine learning model such as a deep neural network. Such a solution, however, is only applicable when users have knowledge about, or at least can estimate, what changes and how the changes will occur. Unfortunately it is not the case in most practical tasks. It becomes even more challenging when we consider the fact that, data in real big data tasks are usually accumulated with time, e.g., instances are being received one by one, like a \textit{stream}. It is impossible to train a machine learning model after we get \textit{all data} at hand as in conventional studies, whereas a more reasonable way is to enable the trained model to be refined/updated according to the newly received data. Unfortunately, it is well known that \textit{catastrophic forgetting} \cite{Pfulb:Gepperth2019} can occur if a trained deep neural network is to be refined with new data only, whereas a frequent re-training based on storing all received data may lead to unbearably huge computational and storage costs. Though there are studies like \textit{continual learning} \cite{Delange:Aljundi:Masana2022} trying to help deep neural networks resist forgetting, many passes scanning over large batches of training data and offline training are generally required, with serious computational and storage concerns on big stream data.

Despite the grand challenges, recently there are considerable research efforts on Open ML. This article will briefly introduce some advances in this line of research, focuses on techniques concerning emerging new classes, decremental/incremental features, changing data distributions, and varied learning objectives. Some theoretical issues will also be discussed.

\section{Emerging New Classes}\label{sec:newclass}

Close-environment machine learning studies generally assume that the class label of any unseen instance $\hat{\bm{x}}$ must be a member of the given label set known in advance, i.e., $\hat{y} \in \mathcal{Y}$. Unfortunately, this does not always hold. For example, consider a forest disease monitor aided by a machine learning model trained with signals sending from sensors deployed in the forest. It is evident that one can hardly enumerate all possible class labels in advance, because some forest diseases can be totally novel, such as those caused by invasive insect pests never encountered in this region before. To be able to handle $\hat{y} \notin \mathcal{Y}$ is a basic requirement for Open ML.

It might be thought that we can artificially generate some virtual training examples for the new classes, just like popular training tricks employed in adversarial deep neural networks. Here, the difficulty lies in the fact that we can hardly imagine what unknown class (called \textit{NewClass} in the following) might occur, whereas training a model accommodating \textit{all possible classes} is impossible or unbearably expensive.

Technically, if \textit{all data} are at hand, especially including the unlabeled instances to be predicted, then handling NewClass can be treated as a special semi-supervised learning \cite{Zhou2018} task, e.g., by establishing the semi-supervised large margin separator corresponding to the tightest contour for each known class, and then regarding unlabeled instances falling outside all contours as NewClass instances \cite{Da:Yu:Zhou2014}. Actually, the distribution of NewClass can be approximated by separating the distribution of known classes for that of the unlabeled data \cite{Zhang:Zhao:Ma:Zhou2020}. Such strategies, however, are not directly applicable when data are accumulated with time.

Consider the following setting of learning with emerging new class. A machine learning model is trained from some initial training data and then deployed to handle unseen instances coming like a stream. For incoming instances of known classes, the trained model should be able to make correct predictions. For incoming instances of unknown class, the model should be able to report that a NewClass instance is encountered; the user can then create a new label for the NewClass. After encountering a few instances of this NewClass, the trained model should be able to be refined/updated such that the NewClass becomes a known class whose incoming instances can be accurately predicted. Ideally, it is desired that the whole process does not require retraining based on storage of \textit{all data} received, since this would be terribly expensive or even infeasible in real big data tasks. Evidently, the above describes an unsupervised/supervised mixing task with human in the loop.

In the first glance, learning with emerging new class seems relevant to \textit{zero-shot learning}, a hot topic in image classification, aiming to classifying visual classes that did not occur in training data set \cite{Socher:Ganjoo:Manning:Ng2013,Xian:Lampert:Schiele:Akata2019,Chen:Geng:Chen:Horrocks2021}. Note that zero-shot learning is assumed to work with \textit{side information}, i.e., external knowledge such as class definitions/descriptions/properties, that can help associate the seen and unseen classes, and thus, it can be treated as a kind of transfer learning \cite{Pan:Yang2009}; in contrast, learning with emerging new class is a general machine learning setting which does not assume such external knowledge. In other words, zero-shot learning assumes that the unseen classes are known, though they did not occur in training data, whereas learning with emerging new class are tackling the grand challenge that the unseen classes are unknown. Thus, approaches for learning with emerging new class can be more general, and can be converted and applied to zero-shot learning.

Classification with \textit{reject} option \cite{Fumera:Roli:Giacinto2000,Bartlett:Wegkamp2008,Geifman:ElYaniv2019} aims to avoiding unconfident predictions that are likely to be incorrect, assuming all classes known in advance. \textit{Open set recognition/classification} \cite{Scheirer:Rocha:Sapkota2013,Geng:Huang:Chen2020} extends reject option to consider the possibility that unknown class may occur in testing phase, with the goal to recognize known classes and reject NewClass. Neither of them attempts to enable the trained model to accommodate NewClass. Some generalized open set recognition studies try to recognize unknown class, by assuming the availability of side information like aforementioned in zero-shot learning \cite{Geng:Huang:Chen2020}, whereas learning with emerging new class is a general machine learning setting which does not assume such external knowledge.

Learning with emerging new class is actually a kind of \textit{incremental learning}, which emphasizes that a trained model only requires slight modification to accommodate new information. There was a long history of studies on incremental learning \cite{Utgoff1989,Syed:Liu:Sung1999,GiraudCarrier2000,He:Chen:Li:Xu2011}, mostly concerning on increment of training examples, i.e., E-IL (example-incremental learning) defined in \cite{Zhou:Chen2002}. Besides E-IL, the other two types of incremental learning defined in \cite{Zhou:Chen2002} are A-IL (attribute-incremental learning) and C-IL (class-incremental learning). A-IL concerns about feature increment, related to what will be discussed in Section~\ref{sec:feature} of this article, though previous studies generally devoted to selecting adequate feature space given all data/features in advance \cite{Ozawa:Toh:Abe:Pang2005,Zhou:Sohn:Lee2012}. C-IL concerns about class increment, related to learning with emerging new class, though previous studies concerned little about the identification of NewClass and generally assumed that the incremental class is known \cite{Masana:Liu:Twardowski2020}.

\textit{Class discovery} \cite{Golub:Slonim:Tamayo:Huard1999,Monti:Tamayo:Mesirov:Golub2003} tries to discover rare classes, as a separate process from class prediction. As mentioned above, learning with emerging new class is an unsupervised/supervised mixing task, while those studies are somewhat relevant to its first phase, mostly unsupervised part.

\begin{figure}[!ht]
\begin{center}
\subfigure[NewClass identification impossible]{\label{fig:fig2a}
   \begin{minipage}[b]{\linewidth}
    \centering
    \includegraphics[width=.55\linewidth]{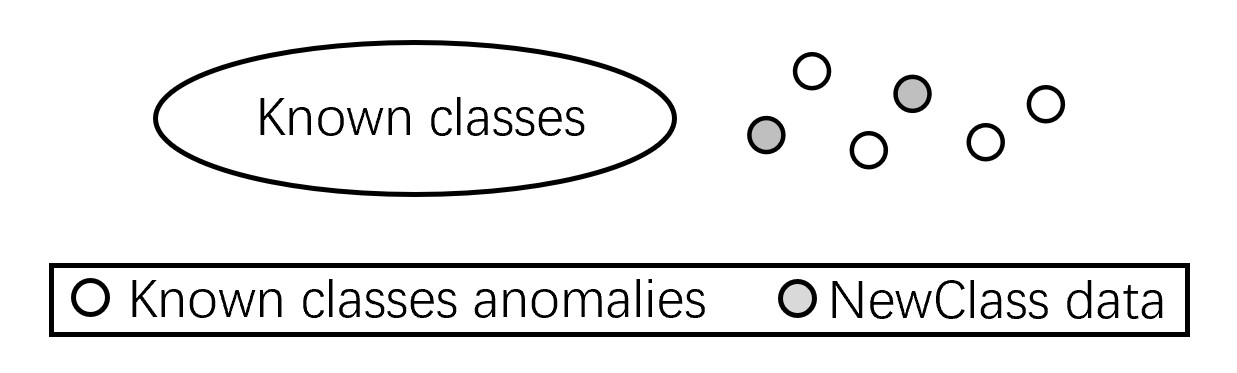}
    \end{minipage}
}
\subfigure[NewClass identification possible]{\label{fig:fig2b}
    \begin{minipage}[b]{\linewidth}
    \centering
    \includegraphics[width=.55\linewidth]{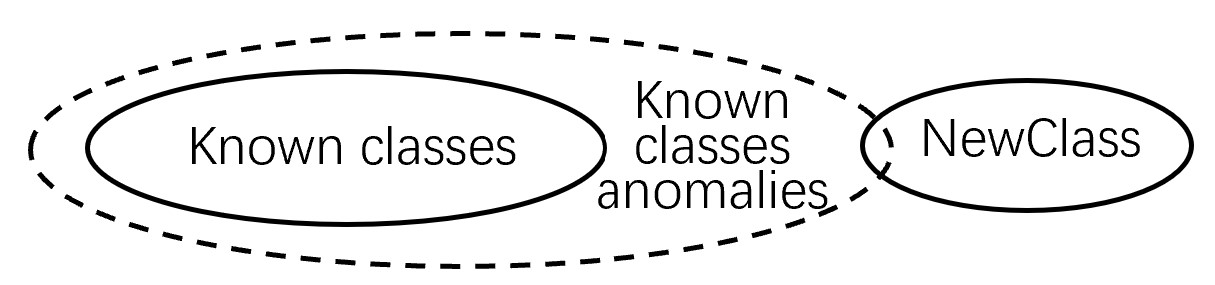}
    \end{minipage}
}
  \caption{NewClass identification not always possible.}\label{fig:NewClass}
\end{center}
\end{figure}

In a general solution \cite{Mu:Ting:Zhou2017} to learning with emerging new class, the first phase, NewClass identification, is realized by anomaly detection. Here, the challenge is to distinguish the NewClass data from anomalies of known classes. In general, this is not always possible; for example, Figure~\ref{fig:NewClass}(a) provides an illustration where the NewClass and anomalies of known classes can hardly be distinguished. Fortunately, in many real tasks it is reasonable to assume that \textit{NewClass instances are more `abnormal' than anomalies of known classes}, as illustrated in Figure~\ref{fig:NewClass}(b). If this does not hold in the original feature space, we can try to identify an adequate feature space by kernel mapping or representation learning. After that, the identification of NewClass instances reduces to anomaly detection from streams, which can be tackled by approaches such as \textit{isolation forest} \cite{Liu:Ting:Zhou:2008}.

Major challenge of the second phase is to refine/update a trained model to accommodate NewClass without sacrificing performance on known classes. For deep neural networks, a retraining based on all data (or at least, on smartly selected subsamples) would be required to avoid catastrophic forgetting, with huge computational and storage costs. It would be ideal to do local refinement only for accommodating NewClass rather than doing global changes that may seriously affect known classes. One solution is to exploit the advantage of tree/forest models through refining only tree leaves involving NewClass in an incremental way \cite{Mu:Ting:Zhou2017}, which does not even need any storage for known class data. Alternatives include techniques that can localize the influence of difference classes such that changes according to NewClass won't significantly affect known classes, such as approaches based on global and local sketching \cite{Mu:Zhu:Du:Lim:Zhou2017}.

If there are more than one new classes, the clustering structure of NewClass data can be exploited \cite{Zhu:Ting:Zhou2017aaai}. Note that there is usually a large gap between the moments when NewClass is detected for the first time and when the model has been refined. To reduce this gap, some efforts have been devoted to enable the model update based on fewer NewClass data \cite{Zhu:Ting:Zhou2017}. \textit{Multi-label learning} with emerging new classes is more challenging because in this scenario the NewClass instances may also hold known class labels, and may even appear in dense regions of known classes, where the key is to detect significant changes in feature combinations and/or label combinations \cite{Zhu:Ting:Zhou2018}. A relevant topic is to examine what known classes are closely related to the NewClass, and an evaluation methodology concerning the mapping from NewClass to known classes has been developed \cite{Faria:Goncalves:Gama2015}.

There are situations where some NewClass instances appeared in training data but misperceived as known class instances, possibly due to insufficiency of feature information. This is even more challenging, with only one very preliminary study having been carried out~\cite{Zhao:Zhang:Zhou2021}.

\section{Decremental/Incremental Features}\label{sec:feature}

Close-environment machine learning studies generally assume that all possible instances, including unseen ones, reside in the same feature space, i.e., $\hat{\bm{x}} \in \mathcal{X}$. Unfortunately, this does not always hold. Taking for example the forest disease monitor mentioned in Section~\ref{sec:newclass}, some sensors could not continue sending signal due to battery exhausted, leading to decremental features, whereas some new sensors can be deployed, leading to incremental features. To be able to handle $\hat{\bm{x}} \in \hat{\mathcal{X}} \neq \mathcal{X}$ is also a requirement for Open ML. Note that in contrast to varied classes where only emerging new class requires special treatment whereas disappeared class can be simply ignored, both decremental and incremental features require attention since feature decrement can lead to seriously downgraded performance.

Consider the following setting of learning with decremental/incremental features. A machine learning model is trained from some initial training data and then deployed to handle unseen data coming like a stream, with decremental and/or incremental features. For incoming testing data, the model should be able to make correct predictions; for incoming additional training data, the model should be able to be refined accordingly. Ideally, it is desired that the whole process does not require retraining based on storage of \textit{all data} received.

\begin{figure}[!ht]
\begin{center}
\subfigure[Phase$_1$ helpless to phase$_2$]{\label{fig:fig3a}
   \begin{minipage}[b]{\linewidth}
    \centering
    \includegraphics[width=0.58\linewidth]{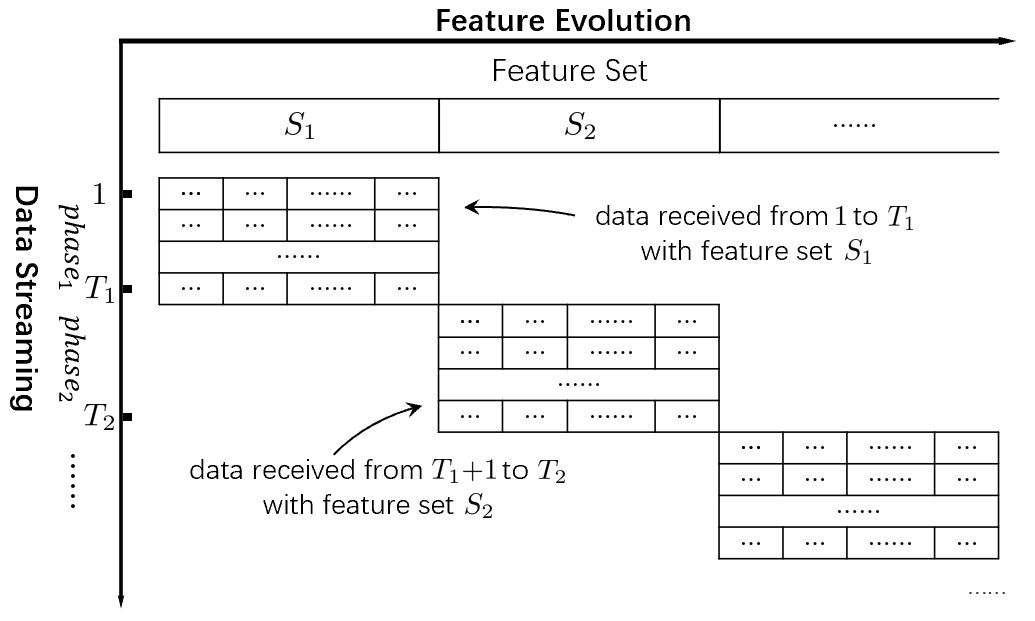}
    \end{minipage}
}
\subfigure[Phase$_1$ helpful to phase$_2$ through survived features]{\label{fig:fig3b}
    \begin{minipage}[b]{\linewidth}
    \centering
    \includegraphics[width=0.58\linewidth]{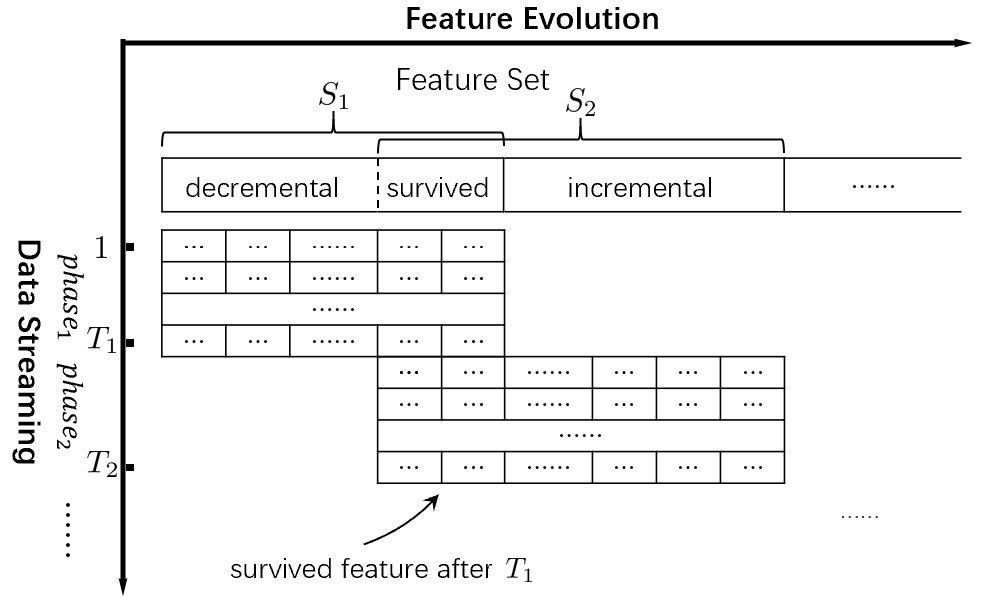}
    \end{minipage}
}
\subfigure[Phase$_1$ helpful to phase$_2$ through shared instances]{\label{fig:fig3c}
    \begin{minipage}[b]{\linewidth}
    \centering
    \includegraphics[width=0.58\linewidth]{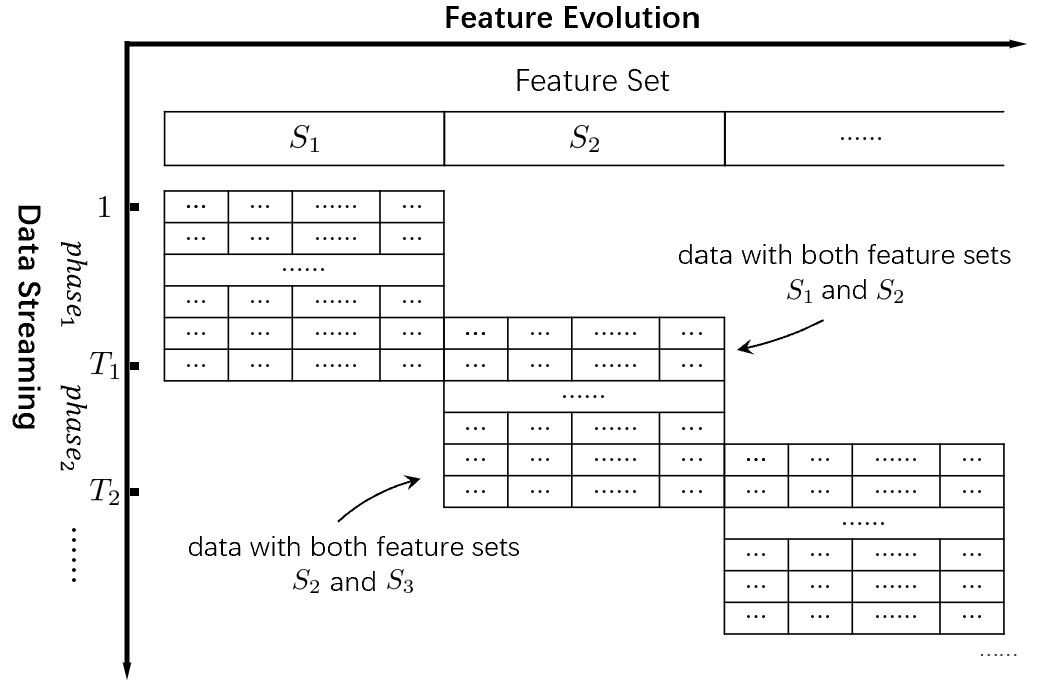}
    \end{minipage}
}
  \caption{Helpless/helpful feature evolution.}\label{fig:features}
\end{center}
\end{figure}

In general, it is not always possible to build a machine learning model which is able to benefit from $\bm{x} \in \mathcal{X}$ for $\hat{\bm{x}} \in \hat{\mathcal{X}} \neq \mathcal{X}$, because machine learning is to learn from experience to improve performance, whereas in most cases there might be little helpful experience from the learning in $\mathcal{X}$ to the learning in $\hat{\mathcal{X}}$ when $\hat{\mathcal{X}} \cap \mathcal{X} = \emptyset$. For example, as illustrated in Figure~\ref{fig:fig3a}, if the feature spaces of phase$_1$ data (i.e., $\{(\bm{x}_1, y_1), \cdots, (\bm{x}_{T_1}, y_{T_1})\}$) and phase$_2$ data (i.e., $\{(\bm{x}_{T_1+1}, y_{T_1+1}), \cdots, (\bm{x}_{T_2}, y_{T_2})\}$) are totally different, then the model trained in phase$_1$ are helpless for phase$_2$, and a new model has to be trained from scratch based on feature set $S_2$ for phase$_2$.

Fortunately, in many real tasks it is often the case that $\hat{\mathcal{X}} \cap \mathcal{X} \neq \emptyset$. In other words, there are features of phase$_1$ survive to be active in phase$_2$ though many other features vanish, as illustrated in Figure~\ref{fig:fig3b}. For example, different sensors may have different battery lifetime, and thus, some old sensors are still working after new sensors being deployed. Formally, in phase$_1$, $\mathcal{X} = \mathcal{X}^{de} \cup \mathcal{X}^{s}$ where $\mathcal{X}^{de}$ and $\mathcal{X}^{s}$ denote the decremental and survived feature sets, respectively; in phase$_2$, $\hat{\mathcal{X}} = \mathcal{X}^{s} \cup \mathcal{X}^{in}$ where $\mathcal{X}^{in}$ denotes the incremental feature set. As $\mathcal{X}^{s}$ is shared in both phases, in addition to training a model$_1$ from $\mathcal{X}$, a model$_2$ based on $\mathcal{X}_{s}$ can be trained in phase$_1$, e.g., by \cite{Hou:Zhou2018}:
\begin{equation}
\begin{split}
&\min\limits_{\bm{w},\bm{w}^{s}} \sum\limits_{i=1}^{T_1}\left(\langle\bm{w},\bm{x}_i\rangle - y_i\right)^{2} + \sum\limits_{i=1}^{T_1}\left(\langle\bm{w}^{s},\bm{x}_i^{s}\rangle - y_i\right)^{2} \\
& \qquad \qquad + \alpha \sum\limits_{i=1}^{T_1}\left(\langle\bm{w},\bm{x}_i\rangle - \langle\bm{w}^{s},\bm{x}_i^{s}\rangle\right)^{2}
+ \gamma \left( \|\bm{w}\|^{2} + \|\bm{w}^{s}\|^{2} \right) \ ,
\end{split}
\end{equation}
where $\langle \cdot, \cdot \rangle$ is inner product, $\bm{w}$ and $\bm{w}^{s}$ are parameters of model$_1$ and model$_2$, respectively, while $\alpha, \gamma > 0$ are the regularization coefficients. Such a process works like `compressing' helpful predictive information from model$_1$ in $\mathcal{X}$ to model$_2$ in $\mathcal{X}^{s}$. Then, in phase$_2$, in addition to the model$_3$ trained based on $\hat{\mathcal{X}}$, the model$_2$ trained in phase$_1$ can still be used. Thus, the prediction in phase$_2$ can be made by combining model$_3$ fed with $\hat{\bm x}$ and model$_2$ fed with the $\mathcal{X}^{s}$ part of $\hat{\bm x}$. In this way, some experience learned from phase$_1$ can be exploited in phase$_2$ through the use of model$_2$.

Interestingly, even when $\hat{\mathcal{X}} \cap \mathcal{X} = \emptyset$, there are cases where it is possible to enable phase$_1$ learning to be helpful to phase$_2$, in particular, when feature increment occurs earlier than feature decrement \cite{Hou:Zhang:Zhou2017}, e.g., new sensors are deployed slightly before old sensors' battery exhausted. As illustrated in Figure~\ref{fig:fig3c}, in this situation there exists a small set of data with both sets of features that can help building a mapping $\psi: \hat{\mathcal{X}} \mapsto \mathcal{X}$. Thus, though $\hat{\bm{x}}$ received in phase$_2$ with features of $\hat{\mathcal{X}}$ only, model$_1$ learned in phase$_1$ can still be exploited by feeding it with $\psi(\hat{\bm{x}})$. Then, phase$_2$ prediction can be made by combining model$_1$ with model$_2$ trained from $\hat{\mathcal{X}}$, either through weighted selection or weighted combination. It has been proved that the cumulative loss of the weighted combination is comparable to the minimum loss between the two models, and the cumulative loss of the weighted selection is comparable to the loss of optimal selection.

The training of these models can be accomplished by online learning techniques such as online gradient descent, and thus, the above strategies can be naturally applied to stream data. It is noticeable that the above strategies can be naturally extended to more phases, and predictions can be made by the combination of multiple models from different feature spaces. Thus, the performance of later phases can be even enhanced by exploiting ensemble learning \cite{Zhou2012}.

Recently, there are studies about learning with feature decrement/increment at unpredictable phase \cite{Hou:Zhang:Zhou2021}, along with data distribution changes \cite{Zhang:Zhao:Jiang:Zhou2020}, 
etc., and applications such as sensor-based activity recognition \cite{Hu:Chen:Peng:Yu2019}.

\section{Changing Data Distributions}\label{sec:changeDistribution}

\begin{figure}[!ht]
\begin{center}
  \includegraphics[width=0.7\linewidth]{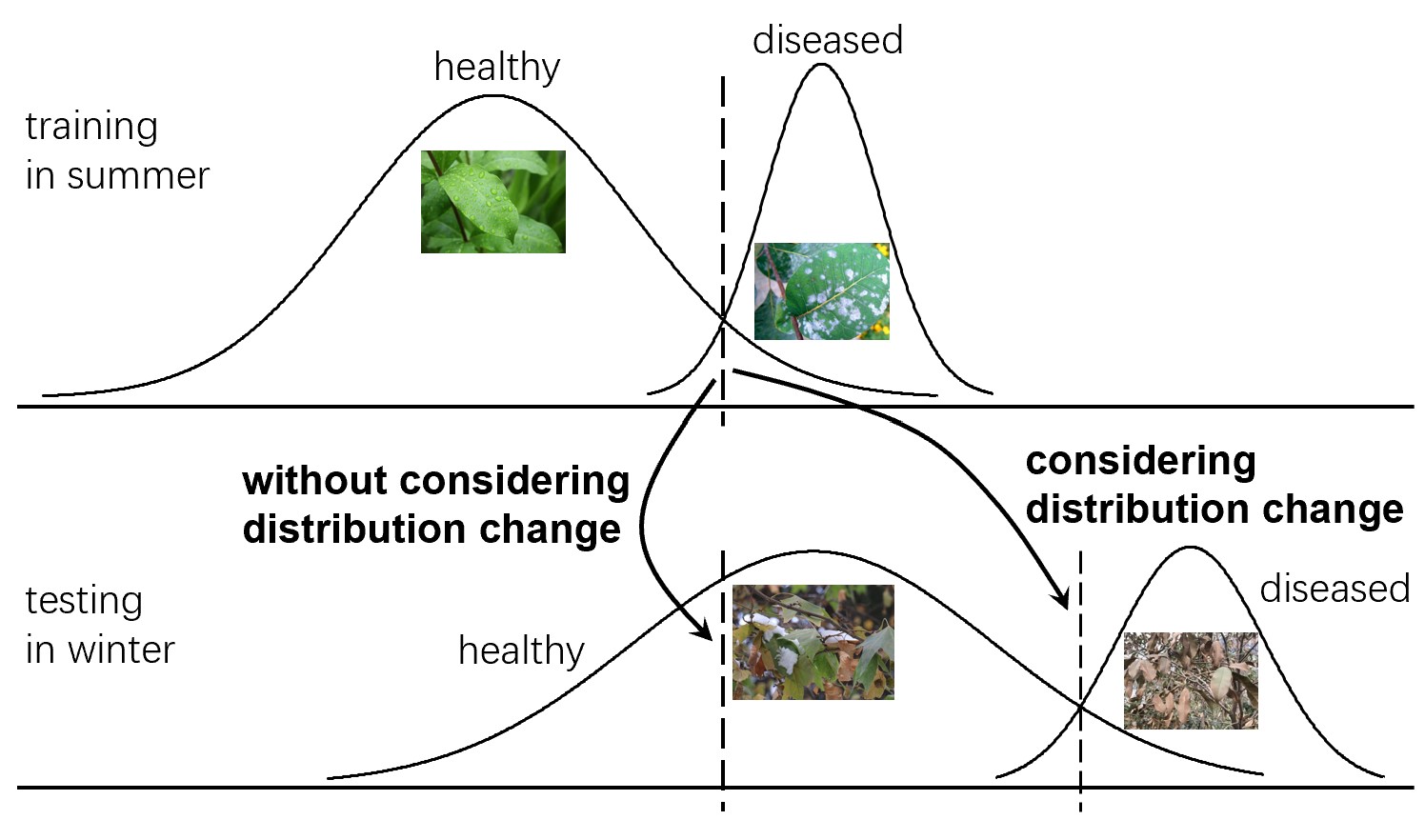}
  \caption{Data distribution change cannot be ignored.}\label{fig:fig4}
\end{center}
\end{figure}

Close-environment machine learning studies generally assume that all data, including both training and testing data, are independent samples from an identical distribution (i.e., \textit{i.i.d.} samples). Unfortunately, this does not always hold. Taking for example the forest disease monitor mentioned in Section~\ref{sec:newclass} again, the model may be built in summer based on sensor signals received in that season, but it is hoped to work well in all seasons. Figure~\ref{fig:fig4} provides an illustration exhibiting that ignoring the data distribution change may lead to seriously downgraded performance.

There have been plentiful studies concerning about training/testing data distribution change. For example, \textit{prior probability shift} and \textit{covariate shift} \cite{Sugiyama:Kawanabe2012} concern about $P_{\text{train}}(y|\bm{x}) = P_{\text{test}}(y|\bm{x})$ but $P_{\text{train}}(y) \neq P_{\text{test}}(y)$ and $P_{\text{train}}(\bm{x}) \neq P_{\text{test}}(\bm{x})$, respectively, whereas \textit{concept drift} \cite{Gama:Zliobaite:Bifet:Pechenizkiy2014} concerns about $P_{\text{train}}(y|\bm{x}) \neq P_{\text{test}}(y|\bm{x})$. Many relevant studies have been conducted under the umbrella of domain adaptation \cite{DaumeIII:Marcu2006,BenDavid:Lu:Luu:Pal2010,Kouw:Loog2021} or transfer learning \cite{Pan:Yang2009}. Note that in steam situation, data distribution change can occur in any phase of the stream, not limited to the testing phase. To be able to handle various kind of data distribution change is an important requirement for Open ML.

In general, learning with changing data distributions is not always possible, e.g., if data distribution can change arbitrarily in every moment without knowledge about how it could change. Fortunately, in many real tasks it is reasonable to assume that \textit{current observation has close relation with recent observations}; in other words, the current instance and the most recent ones are usually from similar or even identical distribution, and \textit{the far the dissimilar}. Thus, we can try to exploit some recent data in the stream to help.

General approaches are often based on \textit{sliding window}, \textit{forgetting}, or \textit{ensemble} mechanisms. Sliding window based approaches hold recent instances and discard old ones falling outside the window, with a fixed or adaptive window size \cite{Klinkenberg:Joachims2000,Kuncheva:Zliobaite2009}. Forgetting based approaches assign a weight to each instance, and downweight old instances according to their age \cite{Koychev2000,Anagnostopoulos:Tasoulis:Adams:Pavlidis2012}. Ensemble \cite{Zhou2012} based approaches add/remove component learners in the ensemble adaptively , and dynamically adjust the weights of learners for incoming instances \cite{Gomes:Barddal:Enembreck2017}.

Most sliding window or ensemble based approaches need to scan data for multiple times. In real big data tasks, it is often hoped that the stream data can be scanned only once and the storage size required by the learning process is independent from the data volume which could not be known before stream ends. Recently, a simple yet effective approach based on forgetting mechanism was proposed to tackle this issue \cite{Zhao:Wang:Xie:Guo:Zhou2021}. The approach does not require prior knowledge about the change, and each instance can be discarded once scanned. Furthermore, inspired by an analysis in control theory \cite{Guo:Ljung:Priouret1993}, a high-probability estimate error analysis based on vector concentration demonstrates that the estimate error decreases until convergence.

Data distribution change can occur in more complicated situations, such as on data with rich structures. There are studies on this issue in multi-instance learning \cite{Foulds:Frank2010}, where the key is to consider both the \textit{bag}-level changes as well as instance-level changes \cite{Zhang:Zhou2017}.

\section{Varied Learning Objectives}

The performance of learning $f: \mathcal{X} \mapsto \mathcal{Y}$ can be measured by a performance measure $M_f$, such as accuracy, F1 measure, and Area under ROC Curve (AUC). Learning towards different objectives may lead to different models with different strengths. A model which is optimal on one measure does not mean that it can also be optimal on other measures. Close-environment machine learning studies generally assume that the $M_f$ which will be used to measure the learning performance should be invariant and known in advance. Unfortunately, this does not always hold. Taking for example the sensor dispatch task, initially many sensors are to be dispatched to pursue a high accuracy of monitoring, whereas after a relatively high accuracy has been achieved, other sensors are to be dispatched to ensure the system continue to work with energy consumption as low as possible. To be able to handle the varied objectives is desired for Open ML.

Learning with varied learning objectives has rarely been studied. Here, the great challenge is to enable a trained machine learning model to switch smoothly from one objective to another, without requiring recollecting data to train a totally new model. There are studies on adapting a trained model to a new objective, based on the observation that many performance measures are relevant \cite{Cortes:Mohri2004,Wu:Zhou2017}; indeed, a large variety of performance measures can be optimized by exploiting nonlinear auxiliary classifiers while keeping high computational efficiency \cite{Li:Tsang:Zhou2013}. This is also relevant to the strategy of \textit{model reuse} \cite{Zhao:Cai:Zhou2020,Wu:Liu:Zhou2019}.

\begin{figure}[!ht]
\begin{center}
  \includegraphics[width=0.55\linewidth]{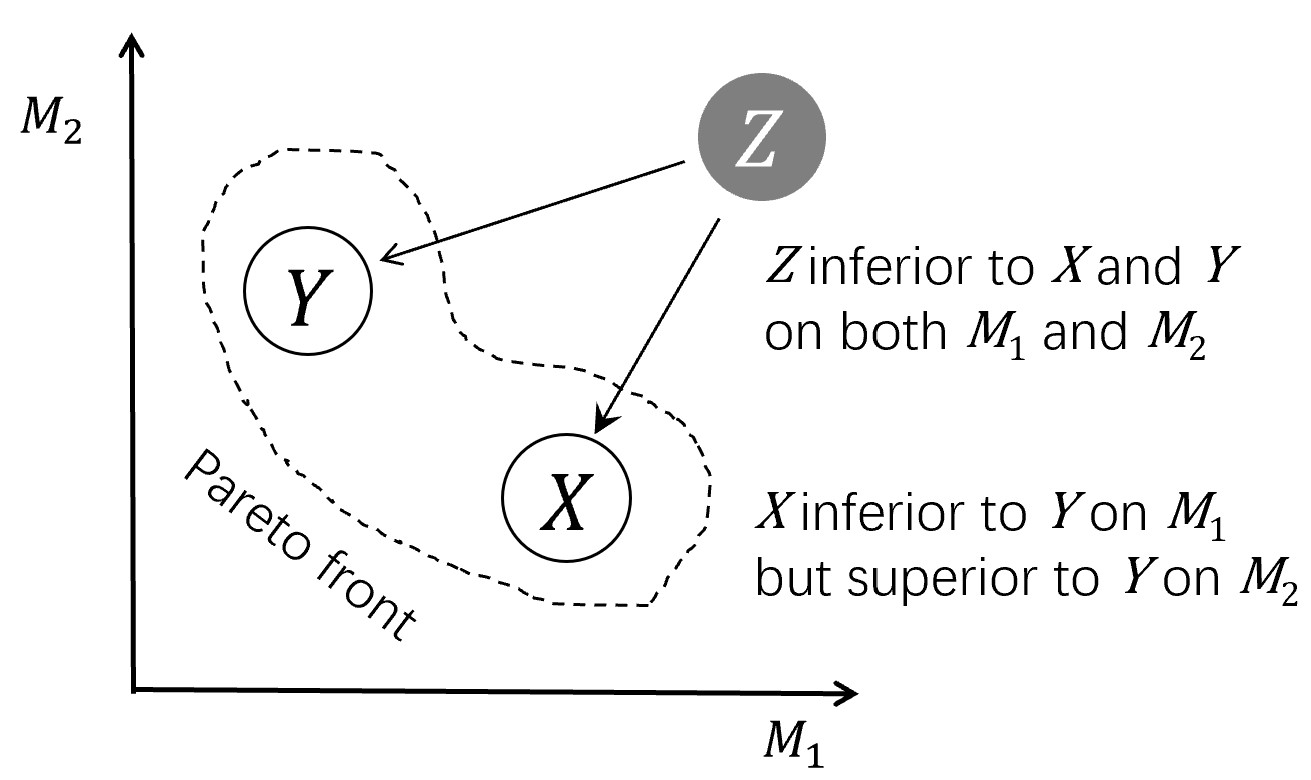}
  \caption{An illustration of pareto front.}\label{fig:fig5}
\end{center}
\end{figure}

In addition to switching from one objective to another, learning with varied learning objectives can also be accomplished by pursuing multiple objectives simultaneously, if these objectives are explicitly known in advance. This resorts to \textit{pareto optimization}. Formally, the goal is to optimize $\min (M_1, M_2, \dots, M_n)$ where $M_i$ are the objectives; the smaller the better. There usually does not exist a single model which is optimal on all objectives; instead, the goal is to seek the \textit{pareto front} consisting of solutions never inferior to other solution on all objectives simultaneously. Figure~\ref{fig:fig5} provides an illustration, where the solutions $X$ and $Y$ are not inferior to any other solution on both objectives simultaneously, so they reside in the pareto front.

Evolutional algorithms, such as genetic algorithms, have been commonly used for pareto optimization in practice, though they are often criticized as they appear to be pure heuristics. It is worth mentioning that recently, there are efforts trying to establish theoretical foundation of evolutionary learning \cite{Zhou:Yu:Qian2019}, i.e., multi-objective machine learning by exploiting evolutionary mechanisms, and it has been shown that the theoretical advances can help guide the design of powerful new algorithms, such as an evolutionary algorithm which is provable to achieve better approximation guarantees than conventional algorithms for the first time.

Besides explicit multiple objectives, implicit multiple objectives also require attention for Open ML. For example, there are situations where users could not express their objectives clearly, but can provide preference feedback like `model$_1$ is better for me than model$_2$'. It has been shown \cite{Ding:Zhou2018} that effective models can be obtained for such kind of implicit objectives by exploiting techniques like \textit{bag of words} \cite{Zhang:Jin:Zhou2010}, assuming that each implicit objective is inherently a kind of combination of element objectives.

\section{Theoretical Issues}

Open ML is a new research direction, and therefore, too many theoretical issues are to be explored.

Among the four threads shown in Figure~\ref{fig:fig1}, current techniques for learning with emerging new classes are mostly based on heuristics \cite{Da:Yu:Zhou2014,Mu:Ting:Zhou2017,Mu:Zhu:Du:Lim:Zhou2017,Zhu:Ting:Zhou2018}. Note that when \textit{all data} are at hand there are some theoretical results, e.g., when NewClasses exist in unlabeled data \cite{Zhang:Zhao:Ma:Zhou2020,Liu:Garrepalli:Hendrycks:Fern2022}; however, these results are not directly applicable when data are accumulated with time, where NewClass \textit{emerging} in stream. There are some theoretical analyses on the proposed algorithms for learning with decremental/incremental features \cite{Hou:Zhou2018,Hou:Zhang:Zhou2017,Hou:Zhang:Zhou2021}, 
but a thorough theoretical study is lacking. Learning with multi-objectives using evolutional mechanisms has its theoretical foundation being established \cite{Zhou:Yu:Qian2019}, but the varied learning objective issue as a whole is underexplored yet. Learning with changing distributions is with relatively more theoretical studies. For example, concept drift has a long thread of theoretical exploration \cite{Helmbold:Long1994,Crammer:Mansour:EvenDar2010,Mohri:Medina2012}, and some algorithms were proposed with theoretical analyses, from view of mistake and loss bounds \cite{Kolter:Maloof2005}, stability analysis \cite{Harel:Mannor:ElYaniv2014}, generalization and regret analysis \cite{Zhao:Cai:Zhou2020}, etc. There are also theoretical studies about relaxing the \textit{i.i.d.} assumption \cite{Mohri:Rostamizadeh2008,Pentina:Lampert2015,Gao:Niu:Zhou2016}.

Open ML is challenging mostly because we can hardly know what changes and how the changes will occur in advance. This is quite different from typical scenarios handled by reinforcement learning \cite{Sutton:Barto2018,Majid:Saaybi:Rietbergen2021} where a learner interacts with the environment to explore the problem space. Once changes concerned in Open ML occur, previous exploration efforts of the reinforcement learner may become invalid since the problem space is altered by the changes. There are studies about adapting a reinforcement learner to a changed environment \cite{Zhang:Yu:Zhou2018,Chen:Yu:Da:Tan2018}, but the changes should not occur frequently or continuously.

\begin{figure}[!ht]
\begin{center}
  \includegraphics[width=0.55\linewidth]{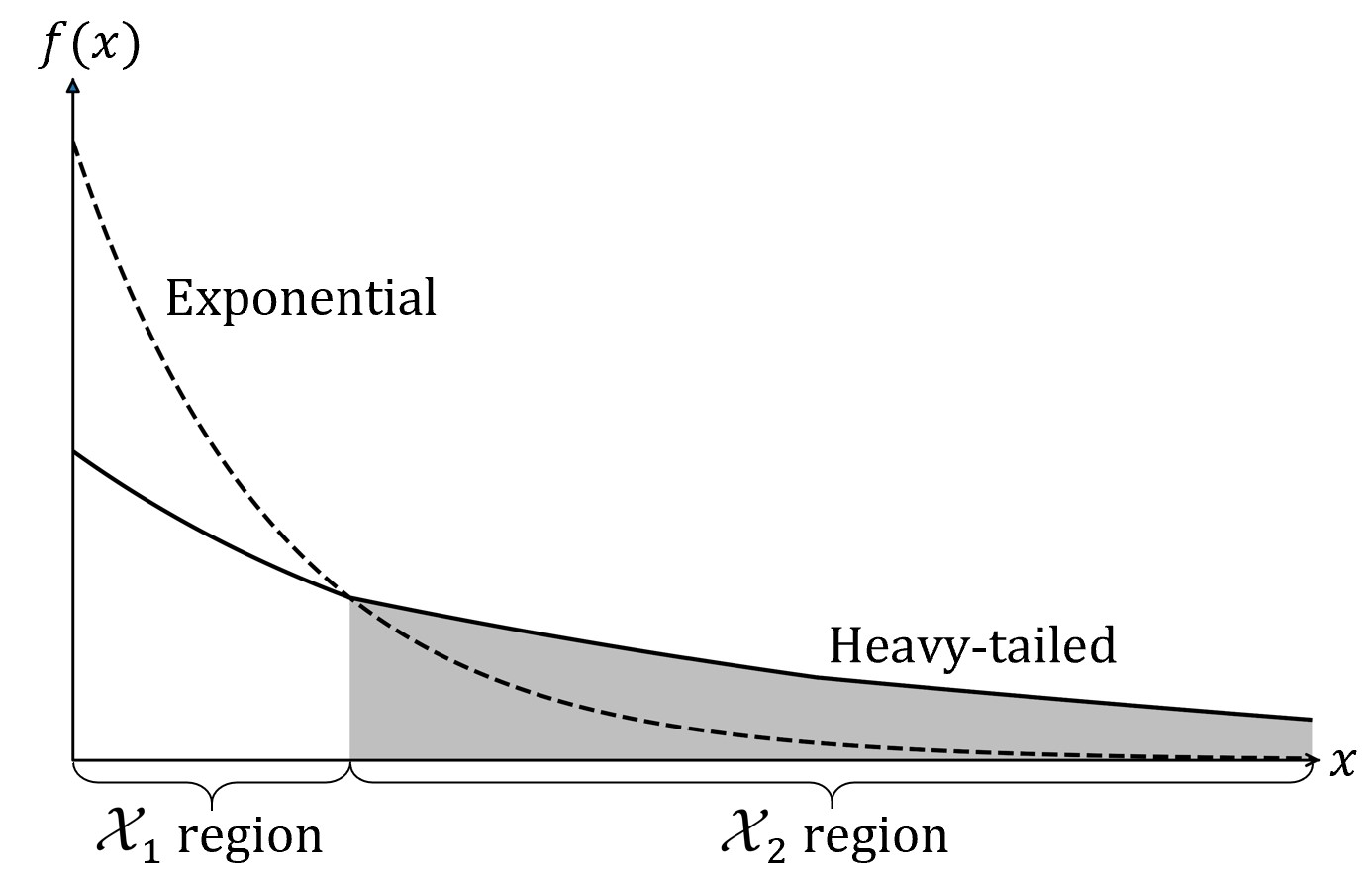}
  \caption{An illustration of heavy-tailed distribution.}\label{fig:fig6}
\end{center}
\end{figure}

Technically, in Open ML one does not have data reflecting unknown changes in the initial training set, while adequate model update must be conducted after receiving a few instances upon changes occur as soon as possible. From this aspect, Open ML is somewhat relevant to \textit{weakly supervised learning} \cite{Zhou2018}. However, in contrast to close-environment studies that emphasize on the \textit{majority} examples and thus generally assume \textit{normal distribution}, in Open ML the \textit{minority} examples or even those that have never been observed are much more important, though at the meantime a good performance on the majority is still demanded. Thus, instead of normal distribution, it would be more favorable to consider \textit{heavy-tailed distributions} (especially \textit{fat-tailed distributions} where very rare events may cause extremely large losses rather than commonly studied \textit{long-tailed distributions}) where the tails are not exponentially bounded, as illustrated in Figure~\ref{fig:fig6}.

Evidently, one hopes the learned model $h(\bm{x})$ satisfies
\begin{equation}\label{eq:def}
P(E_i(h) \leq \epsilon_i) \geq 1-\delta_i \ ,
\end{equation}
where $E_i(h) = P_{\bm{x} \in \mathcal{X}_i} \left(h_i(\bm{x}) \neq y\right)$, $i \in \{1,2\}$, $y$ is ground-truth output of $\bm{x}$, $0 < \epsilon_1 \le \epsilon_2 \le \epsilon$,  $\delta_1, \delta_2 <1$. Intuitively, this explains that the desired model should achieve excellent performance in $\mathcal{X}_1$ region in Figure~\ref{fig:fig6} (i.e., the error should be smaller than $\epsilon_1$ with a high probability), and satisfactory performance in $\mathcal{X}_2$ region (i.e., the error $\epsilon_2$
must not be larger than $\epsilon$ though it can be larger than $\epsilon_1$). The rigid threshold $\epsilon$ is to ensure that the worst performance can be bearable to user no matter what changes occur. This is relevant to \textit{safe learning} \cite{Li:Zhou2015} in weakly supervised scenario, and the principle \textit{optimizing the worst-case performance after achieving a good average performance} can be helpful. Consequently, the total error is
\begin{equation}
E = E_1(h) + \gamma E_2(h) \ ,
\end{equation}
where $\gamma$ is the coefficient to trade off $\mathcal{X}_1$ and $\mathcal{X}_2$ regions, and can be set by user according to relative importance of these regions; $E$ is bounded by $(1+\gamma)\epsilon$ according to Eq.~\ref{eq:def}. The above understanding offers a perspective to regard the $\mathcal{X}_2$ region as a regularization force to the learning in $\mathcal{X}_1$ region.

Typical heavy-tailed distributions include Pareto distribution, Cauchy distribution, etc. When they are assumed instead of the commonly used normal distribution, new challenges arise. For example, the Central Limit Theorem does not hold, and frequent sample statistics, such as the popularly used sample mean and variance, would be misleading (i.e., they can be very different from population mean and variance). These issues must be considered in Open ML. For example, if the input and output spaces are heavy-tailed, empirical risk minimization becomes invalid, since empirical risk is no longer a good approximation of risk \cite{Catoni2012}. This poses problem for learning algorithms, even for simple L1-regression \cite{Zhang:Zhou2018}.

Considering data accumulated with time, performance measure requires attention. Here, the concern is that no matter what changes will occur, the learning process is running as online learning \cite{CesaBianchi:Lugosi2006,ShalevShwartz2011}. In contrast to close-environment studies that assume stationary online setting, Open ML pays attention to non-stationary online setting. As a consequence, rather than static regret which measures the performance by the cumulative loss of the learner against that of the best constant point chosen in hindsight, general dynamic regret \cite{Zinkevich2003} which compares the cumulative loss of the learner against any sequence of comparators is more reasonable. Optimal results have been reported recently on online convex optimization with various mechanisms \cite{Zhang:Lu:Zhou2018,Zhao:Zhang:Zhang:Zhou2020,Zhao:Wang:Zhou2022} 
and bandit convex optimization \cite{Zhao:Wang:Zhang:Zhou2021}. A nearly minimax optimal solution to non-stationary linear bandits under mild condition has been reported through a simple yet effective \textit{restart} mechanism \cite{Zhao:Zhang:Jiang:Zhou2020} with a scheduling scheme~\cite{Wei:Luo2021}, which is more friendly to resource-constrained learning tasks than sliding window or forgetting mechanisms.

Open ML is also related to learning with noisy data, for which there are many theoretical studies, e.g., \cite{Angluin:Laird1988,Blum:Kalai:2003,Natarajan:Dhillon:Ravikumar:Tewari2013,Gao:Wang:Li:Zhou2016,Gao:Zhang:Yang:Zhou2021}.
Note that, in contrast to close-environment studies where noises can be simply depressed by techniques such as smoothing, important signals in Open ML might be hidden in signals that are regarded as noise, and rare important events might be depressed by oversimplified smoothing.

\section{Conclusion}

This article briefly introduces some research advances in open environment machine learning. It can hardly be a thorough review of all relevant work, and is mostly a brief summary of author and his colleagues's exploration along this direction, emphasizing on general principles and strategies rather than specific learning algorithms. Many strategies and ideas mentioned in this article can be realized with various learning techniques, possibly with different strengths to be explored in future. Note that the varied issues are discussed separately in this article, while in real practice they often occur simultaneously. It is fundamentally important to enable machine learning models to achieve excellent performance in usual case while keeping satisfactory performance no matter what unexpected unfortunate issues occur. This is crucial for achieving robust artificial intelligence \cite{Dietterich2017,Dietterich2019}, and carries desired properties of learnware \cite{Zhou2016}.



\bibliographystyle{plain}
\bibliography{openML}

\end{document}